\documentclass[11pt,a4paper]{article}
\usepackage{acl2017}
\usepackage{times}
\usepackage{latexsym}

\usepackage{amsmath}
\usepackage{amsfonts}
\usepackage{amssymb}
\usepackage{booktabs}
\usepackage{mathtools}
\usepackage[mathcal]{eucal}
\usepackage{subcaption}
\usepackage{setspace}
\usepackage{bm}
\usepackage{inconsolata}

\usepackage{tkz-graph}
\usetikzlibrary{backgrounds}

\newcommand{\argpred}[0]{{argument mining}}
\newcommand{\prop}[0]{{proposition}}
\newcommand{\props}[0]{{propositions}}
\newcommand{\links}[0]{{links}}
\newcommand{\link}[0]{{link}}
\newcommand{\Prop}[0]{{Proposition}}
\newcommand{\Link}[0]{{Link}}
\newcommand{\adq}[0]{AD\textsuperscript{3}}

\newcommand{\typefact}{\footnotesize \textsc{fact}}
\newcommand{\typetestimony}{\footnotesize \textsc{testimony}}
\newcommand{\typevalue}{\footnotesize \textsc{value}}
\newcommand{\typepolicy}{\footnotesize \textsc{policy}}
\newcommand{\typereference}{\footnotesize \textsc{reference}}

\newcommand{\typeclaim}{\footnotesize \textsc{claim}}
\newcommand{\typemc}{\footnotesize \textsc{major claim}}
\newcommand{\typepremise}{\footnotesize\textsc{premise}}

\newcommand{\argmax}[0]{\operatorname{arg\,max}}

\newcommand{\defeq}{\vcentcolon=}

\newcommand{\bidirepr}[1]{\overset{\text{\scriptsize$\leftrightarrow$}}{\vphantom{\bm{b}}\bm{#1}}}

\newcommand{\mlpenc}[1]{\overline{\vphantom{\bm{b}}\bm{#1}}}

\newcommand{\tabitem}{\par\hspace*{\labelsep}\textbullet\hspace*{\labelsep}}

\newcommand{\smallspace}{\addlinespace[0.1em]}
\newcommand{\midspace}{\addlinespace[0.3em]}
\newcommand{\cnt}[1]{{\color{mgray} \scriptsize (#1)}}
\def\quad{\hskip0.8em\relax}

\newcommand{\secref}[1]{Section~\ref{sec:#1}}
\newcommand{\tabref}[1]{Table~\ref{tab:#1}}
\newcommand{\figref}[1]{Figure~\ref{fig:#1}}
\newcommand{\eqnref}[1]{Equation~\ref{eq:#1}}

\definecolor{mgray}{gray}{0.4}
\newcommand{\qspopen}[1]{{\color{mgray}[}\,}
\newcommand{\qspclose}[1]{{\color{mgray}\,]$_{#1}$~}}

\aclfinalcopy
\def\aclpaperid{366}

\title{Argument Mining with Structured SVMs and RNNs}

\author{Vlad Niculae\\
Cornell University \\
{\tt vlad@cs.cornell.edu}\\\And
Joonsuk Park \\
Williams College \\
{\tt jpark@cs.williams.edu}\\\And
Claire Cardie\\
Cornell University\\
{\tt cardie@cs.cornell.edu}}

\date{}

\begin{document}
\maketitle
\begin{abstract}
We propose a novel factor graph model for argument mining, designed for settings 
in which the argumentative relations in a document do not necessarily form a tree
structure. (This is the case in over 20\% of the web comments dataset we
release.) Our model jointly learns elementary unit type classification and
argumentative relation prediction. Moreover, our model supports SVM and RNN
parametrizations, can enforce structure constraints (e.g., transitivity), and
can express dependencies between adjacent relations and propositions.
Our approaches outperform unstructured baselines in both web comments and
argumentative essay datasets.

\end{abstract}

\section{Introduction}

\begin{figure}[t]
\begin{quote}
\footnotesize
\setstretch{1.15}
\qspopen{a}Calling a debtor at work is counter-intuitive;\qspclose{a}
\qspopen{b}if collectors are continuously calling someone at work, other
employees may report it to the debtor's supervisor.\qspclose{b}%
\qspopen{c}Most companies have established rules about receiving or making
personal calls during working hours.\qspclose{c}%
\qspopen{d}If a collector or creditor calls a debtor on his/her cell phone and
is informed that the debtor is at work, the call should be
terminated.\qspclose{d}%
\qspopen{e}No calls to employers should be allowed,\qspclose{e}%
\qspopen{f}as this jeopardizes the debtor's job.\qspclose{f}%
\end{quote}
{\centering \footnotesize
\begin{tikzpicture}[>=latex,line join=bevel,scale=0.45]
  \pgfsetlinewidth{1bp}
\pgfsetcolor{black}
  \draw [->] (76.107bp,71.697bp) .. controls (71.735bp,63.644bp) and (66.443bp,53.894bp)  .. (56.785bp,36.104bp);
  \draw [->] (85.5bp,143.7bp) .. controls (85.5bp,135.98bp) and (85.5bp,126.71bp)  .. (85.5bp,108.1bp);
  \draw [->] (113.63bp,143.88bp) .. controls (128.67bp,134.72bp) and (147.37bp,123.34bp)  .. (172.26bp,108.19bp);
  \draw [->] (58.92bp,143.91bp) .. controls (47.263bp,134.81bp) and (34.815bp,122.49bp)  .. (28.5bp,108.0bp) .. controls (19.768bp,87.957bp) and (26.027bp,63.578bp)  .. (37.892bp,36.018bp);
  \draw [->] (316.5bp,143.7bp) .. controls (316.5bp,135.98bp) and (316.5bp,126.71bp)  .. (316.5bp,108.1bp);
\begin{scope}
  \definecolor{strokecol}{rgb}{0.0,0.0,0.0};
  \pgfsetstrokecolor{strokecol}
  \draw (85.5bp,90.0bp) node {$b$ ({\typevalue})};
\end{scope}
\begin{scope}
  \definecolor{strokecol}{rgb}{0.0,0.0,0.0};
  \pgfsetstrokecolor{strokecol}
  \draw (47.5bp,18.0bp) node {$a$ ({\typevalue})};
\end{scope}
\begin{scope}
  \definecolor{strokecol}{rgb}{0.0,0.0,0.0};
  \pgfsetstrokecolor{strokecol}
  \draw (200.5bp,90.0bp) node {$d$ ({\typepolicy})};
\end{scope}
\begin{scope}
  \definecolor{strokecol}{rgb}{0.0,0.0,0.0};
  \pgfsetstrokecolor{strokecol}
  \draw (85.5bp,162.0bp) node {$c$ ({\typefact})};
\end{scope}
\begin{scope}
  \definecolor{strokecol}{rgb}{0.0,0.0,0.0};
  \pgfsetstrokecolor{strokecol}
  \draw (316.5bp,162.0bp) node {$f$ ({\typevalue})};
\end{scope}
\begin{scope}
  \definecolor{strokecol}{rgb}{0.0,0.0,0.0};
  \pgfsetstrokecolor{strokecol}
  \draw (316.5bp,90.0bp) node {$e$ ({\typepolicy})};
\end{scope}
\end{tikzpicture}

}
\caption{Example annotated CDCP comment.\footnotemark}
\label{fig:ex}
\end{figure}
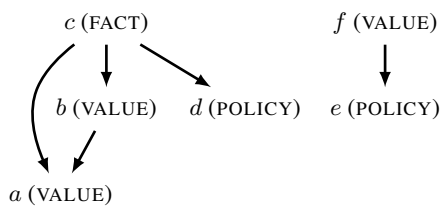

Argument mining consists of the automatic identification of argumentative
structures in documents, a valuable task with applications in policy making,
summarization, and education, among others.  The {\argpred} task includes the
tightly-knit subproblems of classifying {\props} into elementary unit types and
detecting argumentative relations between the elementary units.
The desired output is a document argumentation graph structure, such as the one
in \figref{ex}, where {\props} are denoted by letter subscripts, and the
associated argumentation graph shows their types and support relations between
them.

Most annotation and prediction efforts in {\argpred} have focused on tree or
forest structures \cite{peldstruct,stabcoli}, constraining argument structures
to form one or more trees.  This makes the problem computationally easier by
enabling the use of maximum spanning tree--style parsing approaches.  However,
argumentation {\em in the wild} can be less well-formed.
The argument put forth in \figref{ex}, for instance, consists of two components:
a simple tree structure and a more complex graph structure ($c$ jointly supports
$b$ and $d$).
In this work, we design a flexible and highly expressive structured prediction
model for {\argpred}, jointly learning to classify elementary units (henceforth
{\em \props}) and to identify the argumentative relations between them
(henceforth {\em \links}).  By formulating {\argpred} as inference in a factor
graph~\cite{fg}, our model (described in \secref{learn}) can account for
correlations between the two tasks, can consider second order {\link} structures
(e.g., in \figref{ex}, $c \rightarrow b \rightarrow a$), and can impose
arbitrary constraints (e.g., transitivity).

\footnotetext{We describe
{\prop} types ({\typefact}, %
etc.) in \secref{data}.}

To parametrize our models, we evaluate two alternative directions: linear
structured SVMs \cite{svmstruct}, and recurrent neural networks with
structured loss, extending~\cite{kiperwasser2016simple}. Interestingly, RNNs
perform poorly when trained with classification losses, but become competitive
with the feature-engineered structured SVMs when trained within our proposed
structured learning model.

We evaluate our approach on two {\argpred} datasets.  Firstly, on our new
\textit{Cornell eRulemaking Corpus -- CDCP},\footnote{Dataset available
at \url{http://joonsuk.org}.}
consisting of argument annotations on comments from an eRulemaking discussion
forum, where {\links} don't always form trees (\figref{ex} shows an abridged
example comment, and \secref{data} describes the dataset in more detail).
Secondly, on the UKP argumentative essays v2 (henceforth UKP), where argument
graphs are annotated strictly as multiple trees \cite{stabcoli}.  In both cases,
the results presented in \secref{results} confirm that our models outperform
unstructured baselines.  On UKP, we improve {\link} prediction over the best
reported result in \cite{stabcoli}, which is based on integer linear programming
postprocessing.  For insight into the strengths and weaknesses of the proposed
models, as well as into the differences between SVM and RNN parameterizations,
we perform an error analysis in \secref{erroranalysis}.  To support {\argpred}
research, we also release our Python implementation, {\tt
Marseille}.\footnote{Available at \url{https://github.com/vene/marseille}.}

\section{Related work}
\label{sec:relwork}

Our factor graph formulation draws from ideas previously used independently in
parsing and argument mining.  In particular, maximum spanning tree (MST) methods
for arc-factored dependency parsing have been successfully used by
\newcite{mstparser} and applied to {\argpred} with mixed results by
\newcite{peldstruct}. As they are not designed for the task, MST parsers cannot
directly handle {\prop} classification or model the correlation between {\prop}
and {\link} prediction---a limitation our model addresses. Using RNN features in
an MST parser with a structured loss was proposed by
\newcite{kiperwasser2016simple}; their model can be seen as a particular case of
our factor graph approach, limited to {\link} prediction with a tree structure
constraint. Our models support multi-task learning for {\prop} classification,
parameterizing adjacent {\links} with higher-order structures (e.g., $c
\rightarrow b \rightarrow a$) and enforcing arbitrary constraints on the {\link}
structure, not limited to trees.  Such higher-order structures and logic
constraints have been successfully used for dependency and semantic parsing by
\newcite{turboparser} and \newcite{turbosemanticparser}; to our knowledge we are
the first to apply them to {\argpred}, as well as the first to parametrize them
with neural networks.  \newcite{stabcoli} used an integer linear program to
combine the output of independent {\prop} and {\link} classifiers using a
hand-crafted scoring formula, an approach similar to our baseline.  Our factor
graph method can combine the two tasks in a more principled way, as it fully
learns the correlation between the two tasks without relying on hand-crafted
scoring, and therefore can readily be applied to other argumentation datasets.
Furthermore, our model can enforce the tree structure constraint, required on
the UKP dataset, using MST % inference rather than the exponential number of
cycle constraints used by \newcite{stabcoli}, thanks to the {\adq} inference
algorithm \cite{martins2015ad3}.

Sequence tagging has been applied to the related structured tasks of {\prop}
identification and classification \cite{stabcoli, ivanweb, parkcrf}; integrating
such models is an important next step. Meanwhile, a new direction in {\argpred}
explores {\em pointer networks} \cite{argpointer}; a promising method, currently
lacking support for tree structures and domain-specific constraints.

\section{Data}
\label{sec:data}

We release a new argument mining dataset consisting of user comments about rule
proposals regarding Consumer Debt Collection Practices (CDCP) by the Consumer
Financial Protection Bureau collected from an eRulemaking website,
\url{http://regulationroom.org}.

Argumentation structures found in web discussion forums, such as the eRulemaking
one we use, can be more free-form than the ones encountered in controlled,
elicited writing such as \cite{peldstruct}. For this reason, we adopt the model
proposed by~\newcite{parkerule}, which does not constrain {\links} to form tree
structures, but unrestricted directed graphs.  Indeed, over 20\% of the comments
in our dataset exhibit local structures that would not be allowable in a tree.
Possible {\link} types are \textit{reason} and \textit{evidence}, and {\prop}
types are split into five fine-grained categories: {\typepolicy} and
{\typevalue} contain subjective judgements/interpretations, where only the
former specifies a specific course of action to be taken.  On the other hand,
{\typetestimony} and {\typefact} do not contain subjective expressions, the
former being about personal experience, or ``anecdotal." Lastly,
{\typereference} covers URLs and citations, which are used to point to objective
evidence in an online setting.

In comparison, the UKP dataset~\cite{stabcoli} only makes the syntactic
distinction between {\typeclaim}, {\typemc}, and {\typepremise} types, but it
also includes {\em attack} {\links}.  The permissible {\link} structure is
stricter in UKP, with {\links} constrained in annotation to form one or more
disjoint directed trees within each paragraph.  Also, since web arguments are
not necessarily fully developed, our dataset has many argumentative {\props}
that are not in any argumentation relations. In fact, it isn't unusual for
comments to have no argumentative {\links} at all: 28\% of CDCP comments have no
{\links}, unlike UKP, where all essays have complete argument structures.  Such
comments with no {\links} make the problem harder, emphasizing the importance of
capturing the {\em lack} of argumentative support, not only its presence.

\subsection{Annotation results} Each user comment was annotated by two
annotators, who independently annotated the boundaries and types of {\props}, as
well as the {\links} among them.\footnote{The annotators used the GATE
annotation tool~\cite{Cunningham2011a}.} To produce the final corpus, a third
annotator manually resolved the conflicts,\footnote{Inter-annotator agreement
is measured with Krippendorf's $\alpha$~\cite{krippendorff1980content} with
respect to elementary unit type ($\alpha$=64.8\%) and {\links}
($\alpha$=44.1\%). A separate paper describing the dataset is under
preparation.} and two automatic preprocessing steps were applied: we take the
{\link} transitive closure, and we remove a small number of nested {\props}.%
\footnote{When two {\props} overlap, we keep the one that results in losing the
fewest {\links}. For generality, we release the dataset without this
preprocessing, and include code to reproduce it; we believe that handling
nested argumentative units is an important direction for further research.} The
resulting dataset contains 731 comments, consisting of about 3800 sentences
($\approx$4700 {\props}) and 88k words. Out of the 43k possible pairs of
{\props}, {\links} are present between only 1300 (roughly 3\%). 
In comparison, UKP has fewer documents (402), but they are longer, with a total
of 7100 sentences (6100 {\props}) and 147k words.  Since UKP {\links} only occur
within the same paragraph and {\props} not connected to the argument are removed
in a preprocessing step, {\link} prediction is less imbalanced in UKP, with 3800
pairs of propositions being linked out of a total of 22k (17\%).  We reserve a
test set of 150 documents (973 {\props}, 272 {\links}) from CDCP, and use the
provided 80-document test split from UKP (1266 {\props}, 809 {\links}).

\section{Structured learning\\\quad for {\argpred}}
\label{sec:learn}

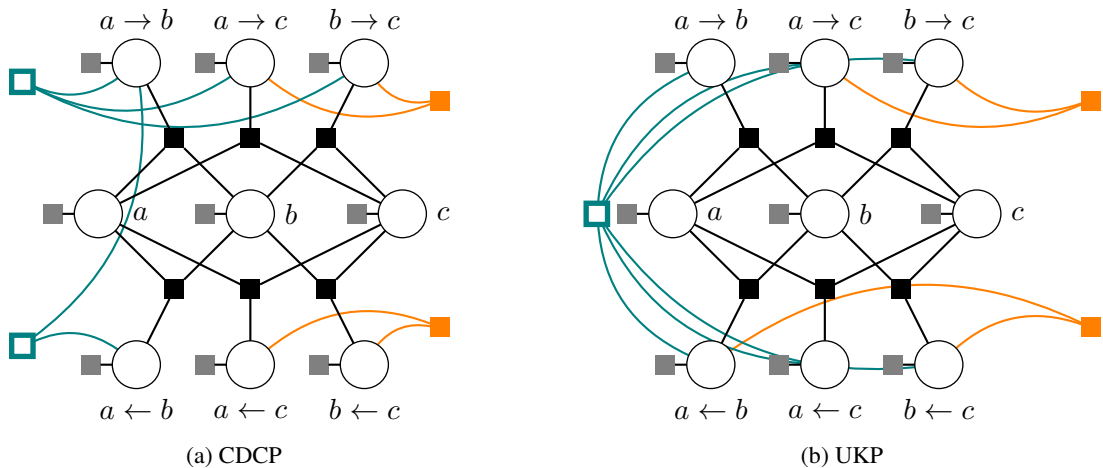
\begin{figure*}[ht]
\begin{minipage}[b]{.5\linewidth}
\centering

\begin{tikzpicture}

    \SetGraphUnit{1}
    \SetUpVertex[Math,Lpos=0,LabelOut]
    \Vertex[L=a,x=0,y=0]{a}
    \Vertex[L=b,x=2,y=0]{b}
    \Vertex[L=c,x=4,y=0]{c}
    \SetUpVertex[Math,Lpos=90,LabelOut]
    \Vertex[L=a\rightarrow{}b,x=0.5,y=2]{ab}
    \Vertex[L=b\rightarrow{}c,x=3.5,y=2]{bc}
    \Vertex[L=a\rightarrow{}c,x=2,y=2]{ac}

    \SetUpVertex[Math,Lpos=-90,LabelOut]
    \Vertex[L=a\leftarrow{}b,x=0.5,y=-2]{ba}
    \Vertex[L=b\leftarrow{}c,x=3.5,y=-2]{cb}
    \Vertex[L=a\leftarrow{}c,x=2,y=-2,Ldist=0.095cm]{ca}
    
    \SetUpVertex[NoLabel]
    \tikzset{VertexStyle/.style={shape=rectangle,fill=black}}
    \Vertex[x=1,y=1]{fab}
    \Vertex[x=3,y=1]{fbc}
    \Vertex[x=2,y=1]{fac}
    
    \Vertex[x=1,y=-1]{fba}
    \Vertex[x=3,y=-1]{fcb}
    \Vertex[x=2,y=-1]{fca}
    \Edge(a)(fab)
    \Edge(b)(fab)
    \Edge(ab)(fab)
    \Edge(a)(fac)
    \Edge(c)(fac)
    \Edge(ac)(fac)
    \Edge(c)(fbc)
    \Edge(b)(fbc)
    \Edge(bc)(fbc)
    \Edge(a)(fba)
    \Edge(b)(fba)
    \Edge(ba)(fba)
    \Edge(a)(fca)
    \Edge(c)(fca)
    \Edge(ca)(fca)
    \Edge(c)(fcb)
    \Edge(b)(fcb)
    \Edge(cb)(fcb)

    \tikzset{VertexStyle/.style={shape=rectangle,fill=gray}}
    \Vertex[x=-0.6,y=0]{fa}
    \Edge(a)(fa)
    \Vertex[x=1.4,y=0]{fb}
    \Edge(b)(fb)
    \Vertex[x=3.4,y=0]{fc}
    \Edge(c)(fc)
    
    \Vertex[x=-0.1,y=2]{fabu}
    \Edge(ab)(fabu)
    \Vertex[x=2.9,y=2]{fbcu}
    \Edge(bc)(fbcu)
    \Vertex[x=1.4,y=2]{facu}
    \Edge(ac)(facu)

    \Vertex[x=-0.1,y=-2]{fbau}
    \Edge(ba)(fbau)
    \Vertex[x=2.9,y=-2]{fcbu}
    \Edge(cb)(fcbu)
    \Vertex[x=1.4,y=-2]{fcau}
    \Edge(ca)(fcau)

    \begin{scope}[on background layer]
    \tikzset{VertexStyle/.style={shape=rectangle,fill=orange}}
    \tikzset{EdgeStyle/.style={color=orange,bend left}}
    \Vertex[x=4.5,y=1.5]{abccop}
    \Vertex[x=4.5,y=-1.5]{cabsib}
    \Edge(abccop)(ac)
    \Edge(abccop)(bc)
    \tikzset{EdgeStyle/.style={color=orange,bend right}}
    \Edge(cabsib)(ca)
    \Edge(cabsib)(cb)
    
    \SetVertexSimple[MinSize = 8pt,
                     LineWidth = 2pt,
                     LineColor = teal,%
                     FillColor = white!60]
    \tikzset{VertexStyle/.append style={shape=rectangle}}
    \Vertex[x=-1,y=1.75]{abc}
    \Vertex[x=-1,y=-1.75]{abba}
    
    \tikzset{EdgeStyle/.style={color=teal,bend left}}
    \Edge(ab)(abc)
    \Edge(ac)(abc)
    \Edge(bc)(abc)
    
    \Edge(ab)(abba)
    \Edge(abba)(ba)
    
    \end{scope}
    
\end{tikzpicture}

\subcaption{CDCP}\label{fig:1a}
\end{minipage}%
\begin{minipage}[b]{.5\linewidth}
\centering

\begin{tikzpicture}

    \SetGraphUnit{1}
    \SetUpVertex[Math,Lpos=0,LabelOut]
    \Vertex[L=a,x=0,y=0]{a}
    \Vertex[L=b,x=2,y=0]{b}
    \Vertex[L=c,x=4,y=0]{c}
    \SetUpVertex[Math,Lpos=90,LabelOut]
    \Vertex[L=a\rightarrow{}b,x=0.5,y=2]{ab}
    \Vertex[L=b\rightarrow{}c,x=3.5,y=2]{bc}
    \Vertex[L=a\rightarrow{}c,x=2,y=2]{ac}

    \SetUpVertex[Math,Lpos=-90,LabelOut]
    \Vertex[L=a\leftarrow{}b,x=0.5,y=-2]{ba}
    \Vertex[L=b\leftarrow{}c,x=3.5,y=-2]{cb}
    \Vertex[L=a\leftarrow{}c,x=2,y=-2,Ldist=0.095cm]{ca}

    \SetUpVertex[NoLabel]
    \tikzset{VertexStyle/.style={shape=rectangle,fill=black}}
    \Vertex[x=1,y=1]{fab}
    \Vertex[x=3,y=1]{fbc}
    \Vertex[x=2,y=1]{fac}
    
    \Vertex[x=1,y=-1]{fba}
    \Vertex[x=3,y=-1]{fcb}
    \Vertex[x=2,y=-1]{fca}
    \Edge(a)(fab)
    \Edge(b)(fab)
    \Edge(ab)(fab)
    \Edge(a)(fac)
    \Edge(c)(fac)
    \Edge(ac)(fac)
    \Edge(c)(fbc)
    \Edge(b)(fbc)
    \Edge(bc)(fbc)
    \Edge(a)(fba)
    \Edge(b)(fba)
    \Edge(ba)(fba)
    \Edge(a)(fca)
    \Edge(c)(fca)
    \Edge(ca)(fca)
    \Edge(c)(fcb)
    \Edge(b)(fcb)
    \Edge(cb)(fcb)

    \tikzset{VertexStyle/.style={shape=rectangle,fill=gray}}
    \Vertex[x=-0.6,y=0]{fa}
    \Edge(a)(fa)
    \Vertex[x=1.4,y=0]{fb}
    \Edge(b)(fb)
    \Vertex[x=3.4,y=0]{fc}
    \Edge(c)(fc)
    
    \Vertex[x=-0.1,y=2]{fabu}
    \Edge(ab)(fabu)
    \Vertex[x=2.9,y=2]{fbcu}
    \Edge(bc)(fbcu)
    \Vertex[x=1.4,y=2]{facu}
    \Edge(ac)(facu)

    \Vertex[x=-0.1,y=-2]{fbau}
    \Edge(ba)(fbau)
    \Vertex[x=2.9,y=-2]{fcbu}
    \Edge(cb)(fcbu)
    \Vertex[x=1.4,y=-2]{fcau}
    \Edge(ca)(fcau)

    \begin{scope}[on background layer]
    \tikzset{VertexStyle/.style={shape=rectangle,fill=orange}}
    \tikzset{EdgeStyle/.style={color=orange,bend left}}
    \Vertex[x=5.5,y=-1.5]{gramp}
    \Vertex[x=5.5,y=1.5]{abccop}
    \Edge(abccop)(ac)
    \Edge(abccop)(bc)
    \tikzset{EdgeStyle/.style={color=orange,bend right}}
    \Edge(gramp)(ba)
    \Edge(gramp)(cb)
    \SetVertexSimple[MinSize = 8pt,
                     LineWidth = 2pt,
                     LineColor = teal,%
                     FillColor = white!60]
    \tikzset{VertexStyle/.append style={shape=rectangle}}
    \tikzset{EdgeStyle/.style={color=teal,bend left}}
    \Vertex[x=-1,y=0]{tree}
    \Edge(tree)(ab)
    \Edge(tree)(ac)
    \Edge(tree)(bc)
    \tikzset{EdgeStyle/.style={color=teal,bend right}}
    \Edge(tree)(ba)
    \Edge(tree)(cb)
    \Edge(tree)(ca)
    \end{scope}
    
\end{tikzpicture}

\subcaption{UKP}\label{fig:1b}
\end{minipage}

\caption{Factor graphs for a document with three propositions ($a, b, c$) and the six possible edges between them, and some of the factors used, illustrating differences and similarities between our models for the two datasets. 
Unary factors are light gray; compatibility factors are black.
Factors not part of the basic model have curved edges:
higher-order factors are orange and on the right;
{\link} structure factors are hollow, as that they don't have any parameters.
Strict constraint factors are omitted for simplicity.
}
\label{fig:fgs}
\end{figure*}

\subsection{Preliminaries}
Binary and multi-class classification have been applied with some success to
{\prop} and {\link} prediction separately, but we seek a way to jointly learn
the {\argpred} problem at the {\em document} level, to better model contextual
dependencies and constraints.  We therefore turn to {\em structured learning}, a
framework that provides the desired level of expressivity.

In general, learning from a dataset of documents $x_i \in \mathcal{X}$ and their
associated labels $y_i \in \mathcal{Y}$ involves seeking model parameters
$\bm{w}$ that can ``pick out'' the best label under a scoring function $f$:
\begin{equation} \hat{y} \defeq \argmax_{y \in \mathcal{Y}} f (x, y; \bm{w}).
\label{eq:decode} \end{equation}
Unlike classification or regression, where $\mathcal{X}$ is usually a feature
space $\mathbb{R}^d$ and $\mathcal{Y} \subseteq \mathbb{R}$ (e.g., we predict an
integer class index or a probability), in structured learning, more complex
inputs and outputs are allowed.  This makes the $\argmax$ in \eqnref{decode}
impossible to evaluate by enumeration, so it is desirable to find models that
decompose over smaller units and dependencies between them; for instance, as
{\em factor graphs}.  In this section, we give a factor graph description of our
proposed structured model for {\argpred}.

\subsection{Model description} An input document is a string of words with
{\prop} offsets delimited. We denote the {\props} in a document by $\{a, b, c,
...\}$ and the possible directed {\link} between $a$ and $b$ as $a \rightarrow
b$.  The argument structure we seek to predict consists of the type of each
proposition $y_a \in \mathcal{P}$ and a binary label for each link $y_{a
\rightarrow b} \in \mathcal{R} = \{\mathtt{on}, \mathtt{off}\}$.\footnote{For
simplicity and comparability, we follow \newcite{stabcoli} in using binary
{\link} labels even if {\links} could be of different types. This can be
addressed in our model by incorporating ``labeled link'' factors.} The possible
{\prop} types $\mathcal{P}$ differ for the two datasets; such differences are
documented in \tabref{differences}.  As we describe the variables and factors
constituting a document's factor graph, we shall refer to \figref{fgs} for
illustration.

\paragraph{Unary potentials.} Each {\prop} $a$ and each {\link} $a \rightarrow
b$ has a corresponding random variable in the factor graph (the circles in
\figref{fgs}).  To encode the model's belief in each possible value for
these variables, we parametrize the {\em unary factors} (gray boxes in
\figref{fgs}) with unary potentials: $\bm{\phi}(a) \in
\mathbb{R}^{|\mathcal{P}|}$ is a score of $y_a$ for each possible {\prop}
type. Similarly, {\link} unary potentials $\bm{\phi}(a \rightarrow
b)\in\mathbb{R}^{|\mathcal{R}|}$ are scores for $y_{a\rightarrow b}$ being
$\mathtt{on}$/$\mathtt{off}$.  Without any other factors, this  would amount
to independent classifiers for each task.

\paragraph{Compatibility factors.} 
For every possible {\link} $a \rightarrow b$, the variables
$(a, b, a \rightarrow b)$ are bound by a dense factor scoring
their joint assignment (the black boxes in \figref{fgs}).
Such a factor could automatically learn to encourage {\links} from compatible
types (e.g., from {\typetestimony} to {\typepolicy})
or discourage {\links} between less compatible ones (e.g., from {\typefact} to
{\typetestimony}).
In the simplest form, this factor would be parametrized as a 
tensor $\bm{\mathcal{T}} \in \mathbb{R}^{|\mathcal{P}| \times |\mathcal{P}|
\times |\mathcal{R}|}$, with $t_{ijk}$ retaining the score of a source
proposition of type $i$ to be ($k=\mathtt{on}$) or not to be ($k=\mathtt{off}$)
in a {\link} with a proposition of type $j$. For more flexibility, we
parametrize this factor with {\bf compatibility features} depending only on
simple structure: $\bm{t}_{ijk}$ becomes a vector, and the score of
configuration $(i, j, k)$ is given by $\bm{v}_{ab}^\top \bm{t}_{ijk}$ where
$\bm{v}_{ab}$  consists of three binary features:

\begin{itemize}
\item {\bf bias:} a constant value of $1$, allowing $\bm{\mathcal{T}}$ to learn a base score for a label configuration $(i, j, k)$, as in the simple form above,
\item {\bf adjacency:} when there are no other {\props} between the source and the target,
\item {\bf order:} when the source precedes the target.
\end{itemize}

\paragraph{Second order factors.} Local argumentation graph structures such as
$a \rightarrow b \rightarrow c$ might be modeled better together rather than
through separate link factors for $a \rightarrow b$ and $b \rightarrow c$.  As
in higher-order structured models for semantic and dependency
parsing~\cite{turboparser,turbosemanticparser}, we implement three types of
second order factors:
{\bf grandparent} ($a \rightarrow b \rightarrow c$),
{\bf sibling} ($a \leftarrow b \rightarrow c$), and
{\bf co-parent} ($a \rightarrow b \leftarrow c$).
Not all of these types of factors make sense on all datasets: as sibling
structures cannot exist in directed trees, we don't use sibling factors on UKP.
On CDCP, by transitivity, every grandparent structure implies a corresponding
sibling, so it is sufficient to parametrize siblings.  This difference between
datasets is emphasized in \figref{fgs}, where one example of each type of factor
is pictured on the right side of the graphs (orange boxes with curved edges): on
CDCP we illustrate a co-parent factor (top right) and a sibling factor (bottom
right), while on UKP we show a co-parent factor (top right) and a grandparent
factor (bottom right).
We call these factors {\em second order} because they involve two {\link}
variables, scoring the joint assignment of both {\links} being \texttt{on}.

\begin{table*}[t]
\small
\begin{tabular}{p{0.137\textwidth}p{.46\textwidth}p{.32\textwidth}}
\toprule
Model part & CDCP dataset & UKP dataset \\
\midrule
{\prop} types&
{\typereference} $\succ$ {\typetestimony} $\succ$ {\typefact} $\succ$
{\typevalue} $\succ$ {\typepolicy} & 
{\typeclaim}, {\typemc}, {\typepremise} \\ \midspace
{\links} & all possible & within each paragraph \\ \midspace
2\textsuperscript{nd} order factors & siblings, co-parents & grandparents, co-parents \\ \midspace
{\link} structure & transitive acyclic:
    \tabitem $ a \rightarrow b \And b \rightarrow c \implies a \rightarrow c$
    \tabitem \textsc{AtMostOne}($a \rightarrow b$, $b \rightarrow a$)
& directed forest:
    \tabitem \textsc{TreeFactor} over each paragraph
    \tabitem zero-potential ``root'' links $a \rightarrow *$
\\ \midspace
strict constraints &
    {\link} source must be as least as objective as the target:
    \par \quad
    $a \rightarrow b \implies a \succeq b$
& {\link} source must be premise:\par \quad
    $a \rightarrow b \implies a =$~{\typepremise}
\\
\bottomrule
\end{tabular}
\caption{Instantiation of model design choices for each dataset.}
\label{tab:differences}
\end{table*}

\paragraph{Valid {\link} structure.} The global structure of argument {\links}
can be further constrained using domain knowledge.  We implement this using
constraint factors; these have no parameters and are denoted by empty boxes in
\figref{fgs}.
In general, well-formed arguments should be cycle-free.  In the UKP dataset,
{\links} form a directed forest and can never cross paragraphs. This particular
constraint can be expressed as a series of {\em tree factors},\footnote{A tree
    factor regards each bound variable as an edge in a graph and assigns
    $-\infty$ scores to configurations that are not valid trees.
For inference, we can use maximum spanning arborescence
algorithms such as Chu-Liu/Edmonds.} one for each paragraph (the factor
connected to all {\link} variables in \figref{fgs}).
In CDCP,  {\links} do not form a tree, but we use logic
constraints to enforce transitivity (top left factor in \figref{fgs}) and to
prevent symmetry (bottom left); the logic formulas implemented by these factors
are described in \tabref{differences}.  Together, the two constraints have the
desirable side effect of preventing cycles.

\paragraph{Strict constraints.} We may include further domain-specific
constraints into the model, to express certain disallowed configurations.  For
instance, proposition types that appear in CDCP data can be ordered by the level
of objectivity \cite{parkerule}, as shown in \tabref{differences}.
In a well-formed argument, we would want to see {\links} from more objective to
equally or less objective {\props}: it's fine to provide {\typefact}
as reason for {\typevalue}, but not
the other way around.  While the training data sometimes violates this
constraint, enforcing it might provide a useful inductive bias.

\paragraph{Inference.} The $\argmax$ in Equation \ref{eq:decode} is a MAP over a
factor graph with cycles and many overlapping factors, including logic factors.
While exact inference methods are generally unavailable, our setting is
perfectly suited for the Alternating Directions Dual Decomposition ({\adq})
algorithm: approximate inference on expressive factor graphs with overlapping
factors, logic constraints, and generic factors (e.g., directed tree factors)
defined through maximization oracles \cite{martins2015ad3}.
When {\adq} returns an integral solution, it is globally optimal, but
when solutions are fractional, several options are available. At test time,
for analysis, we retrieve exact solutions using the branch-and-bound method.
At training time, however, fractional solutions can be used {\em as-is};
this makes better use of each iteration and actually increases the ratio of
integral  solutions in future iterations, as well as at test time, as proven 
by \newcite{meshi2016train}. We also find that after around 15 training
iterations with fractional solutions, over 99\% of inference calls are integral.
\paragraph{Learning.}
We train the models by minimizing the structured hinge loss 
\cite{mmmn}:
\makeatletter
\def\maketag@@@#1{\hbox to \displaywidth{\hss\m@th\normalfont#1}}
\makeatother
\begin{equation}
\sum_{(x, y) \in \mathcal{D}} %
    \max_{y' \in \mathcal{Y}} %
    (f(x, y'; \bm{w}) + \rho(y, y')) - f(x, y; \bm{w}) %
\label{eq:ssvm}
\end{equation}
\noindent where $\rho$ is a configurable misclassification cost.
The $\max$ in Equation~\ref{eq:ssvm} is not the same as the one used for
prediction, in Equation~\ref{eq:decode}. However, when the cost function $\rho$
decomposes over the variables, cost-augmented inference amounts to regular
inference after augmenting the potentials accordingly.
We use a weighted Hamming cost: \[ \rho(y, \hat{y}) \defeq \sum_v \rho(y_v)
\mathbb{I}[y_v = \hat{y}_v] \]
\noindent where $v$ is summed over all variables in a document $\{a\} \cup \{a
\rightarrow b\}$, and $\rho(y_v)$ is a misclassification cost.  We assign
uniform costs $\rho$ to 1 for all mistakes except false-negative {\links}, where
we use higher cost proportional to the class imbalance in the training split,
effectively giving more weight to positive {\links} during training.

\subsection{Argument structure SVM} One option for parameterizing the potentials
of the unary and higher-order factors is with {\em linear models}, using
{\prop}, {\link}, and higher-order features.
This gives birth to a linear structured SVM~\cite{svmstruct}, which, when using
$l_2$ regularization, can be trained efficiently in the dual using the online
block-coordinate Frank-Wolfe algorithm of \newcite{lacoste2013block}, as
implemented in the {\tt pystruct} library \cite{muller2014pystruct}. This
algorithm is more convenient than subgradient methods, as it does not require
tuning a learning rate parameter.  \paragraph{Features.} For unary {\prop} and
{\link} features, we faithfully follow \newcite[Tables 9 and 10]{stabcoli}:
{\prop} features are lexical (unigrams and dependency tuples), structural (token
statistics and {\prop} location), indicators (from hand-crafted lexicons),
contextual, syntactic (subclauses, depth, tense, modal, and POS), probability,
discourse~\cite{discourse}, and average GloVe embeddings \cite{glove}.
{\Link} features are lexical (unigrams), syntactic (POS and productions),
structural (token statistics, {\prop} statistics and location features),
hand-crafted indicators, discourse triples, PMI, and shared noun counts.

Our proposed higher-order factors for grandparent, co-parent, and sibling
structures require features extracted from a {\prop} triplet $a, b, c$.  In
dependency and semantic parsing, higher-order factors capture relationships
between words, so sparse indicator features can be efficiently used.  In our
case, since propositions consist of many words, BOW features may be too noisy
and too dense; so for simplicity we again take a cue from the {\link}-specific
features used by \newcite{stabcoli}.  Our higher-order factor features are: same
sentence indicators (for all 3 and for each pair), {\prop} order (one for each
of the 6 possible orderings), Jaccard similarity (between all 3 and between each
pair), presence of any shared nouns (between all 3 and between each pair), and
shared noun ratios: nouns shared by all 3 divided by total nouns in each {\prop}
and each pair, and shared nouns between each pair with respect to each {\prop}.
Up to vocabulary size difference, our total feature dimensionality is
approximately 7000 for {\props} and 2100 for {\links}. The number of second
order features is 35.

\paragraph{Hyperparameters.}
We pick the SVM regularization parameter
$C \in \{0.001,\linebreak[1] 0.003,\linebreak[1] 0.01,\linebreak[1]
0.03,\linebreak[1] 0.1,\linebreak[1] 0.3\}$ by k-fold cross validation at
document level, optimizing for the average between {\link} and {\prop}
$F_1$ scores.

\subsection{Argument structure RNN} Neural network methods have proven effective
for natural language problems even with minimal-to-no feature engineering.
Inspired by the use of LSTMs~\cite{lstm} for MST dependency parsing by
\newcite{kiperwasser2016simple}, we parametrize the potentials in our factor
graph with an LSTM-based neural network,\footnote{We use the {\tt dynet}
library~\cite{dynet}.} replacing MST inference with the more general {\adq}
algorithm, and using relaxed solutions for training when inference is inexact.

We extract embeddings of all words with a corpus frequency $>1$, initialized
with GloVe word vectors. We use a deep bidirectional LSTM to encode contextual
information, representing a {\prop} $a$ as the average of the LSTM outputs of
its words, henceforth denoted $\bidirepr{a}$.

\paragraph{{\Prop} potentials.} We apply a multi-layer perceptron (MLP) with
rectified linear activations to each proposition, with all layer dimensions
equal except the final output layer, which has size $|\mathcal{P}|$ and is not
passed through any nonlinearities.

\paragraph{{\Link} potentials.} To score a dependency $a\rightarrow b$,
\newcite{kiperwasser2016simple}
pass the concatenation $[\bidirepr{a}; \bidirepr{b}]$ through an MLP.  After
trying this, we found slightly better performance by first passing {\em each}
{\prop} through a slot-specific dense layer
$\bigl(\mlpenc{a} \defeq \sigma_{\text{src}}(\bidirepr{a}), \mlpenc{b} \defeq
\sigma_{\text{trg}}(\bidirepr{b})\bigr)$ followed by a {\em bilinear}
transformation: $$ \phi_{\text{on}}(a \rightarrow b) \defeq
\mlpenc{a}^\top\bm{W} \mlpenc{b} + \bm{w}_{\text{src}}^\top \mlpenc{a} +
\bm{w}_{\text{trg}}^\top \mlpenc{b} + w_0^{\text{(on)}}.$$ 
\noindent Since the bilinear expression returns a scalar, but the {\link}
potentials must have a value for both the $\mathtt{on}$ and $\mathtt{off}$
states, we set the full potential to $\bm{\phi}(a \rightarrow b) \defeq
[\phi_{\text{on}}(a \rightarrow b), w_0^{\text{(off)}}]$ where
$w_0^{\text{(off)}}$ is a learned scalar bias.  We initialize $\bm{W}$ to the
diagonal identity matrix.

\paragraph{Second order potentials.} Grandparent potentials $\phi(a \rightarrow
b \rightarrow c)$ score two adjacent directed edges, in other words three
{\props}.  We again first pass each {\prop} representation through a
slot-specific dense layer.
We implement a multilinear scorer analogously to the {\link} potentials:
\[ \phi(a \rightarrow b \rightarrow c) \defeq
\sum_{i,j,k} \mlpenc{a}_i \mlpenc{b}_j \mlpenc{c}_k w_{ijk} \]
\noindent where  $\bm{\mathcal{W}} = (w)_{ijk}$ is a third-order cube tensor.  To
reduce the large numbers of parameters, we implicitly represent
$\bm{\mathcal{W}}$ as a rank $r$ tensor:
$w_{ijk} = \sum_{s=1}^{r} u^{(1)}_{is} u^{(2)}_{js} u^{(3)}_{ks}$. 
Notably, this model captures only third-order interactions between the
representation of the three propositions. To capture first-order ``bias" terms,
we could include slot-specific linear terms, e.g., $\bm{w}_a ^\top \mlpenc{a}$;
but to further capture quadratic {\em backoff} effects (for instance, if two
{\props} carry a strong signal of being siblings regardless of their parent), we
would require quadratically many parameters. Instead of explicit lower-order
terms, we propose {\em augmenting} $\mlpenc{a}, \mlpenc{b}$, and $\mlpenc{c}$
with a constant feature of $1$, which has approximately the same effect, while
benefiting from the parameter sharing in the low-rank factorization; an effect
described by \newcite{blondelpolynet}.  Siblings and co-parents factors are
similarly parametrized with their own tensors.

\paragraph{Hyperparameters.}  We perform grid search using k-fold document-level
cross-validation, tuning the dropout probability in the dense MLP layers over
$\{0.05, 0.1, 0.15, 0.2, 0.25\}$ and the optimal number of passes over the
training data over $\{10, 25, 50, 75, 100\}$.  We use 2 layers for the LSTM and
the proposition classifier, 128 hidden units in all layers, and a multilinear
decomposition with rank $r=16$, after preliminary CV runs.

\subsection{Baseline models} We compare our proposed models to equivalent
independent unary classifiers. The unary-only version of a structured SVM is an
$l_2$-regularized linear SVM.\footnote{We train our SVM using SAGA \cite{saga}
in \texttt{lightning} \cite{lightning}.  } For the RNN, we compute unary
potentials in the same way as in the structured model, but apply independent
hinge losses at each variable, instead of the global structured hinge loss.
Since the RNN weights are shared, this is a form of multi-task learning.
The baseline predictions can be interpreted as unary potentials, therefore
we can simply round their output to the highest scoring labels, or we can,
alternatively, perform test-time inference, imposing the desired structure. 

\section{Results}
\label{sec:results}

\begin{table*}[ht!]
\centering
    \small
   \begin{tabular}{p{2.65cm} @{\hskip 0.2cm} r r r r r r r r r r r r}
   \toprule 
         & \multicolumn{6}{c}{Baseline}
         & \multicolumn{6}{c}{Structured}\\
     \cmidrule(l{2pt}r{2pt}){2-7} \cmidrule(l{2pt}r{2pt}){8-13} 
         & \multicolumn{3}{c}{SVM}
         & \multicolumn{3}{c}{RNN}
         & \multicolumn{3}{c}{SVM}
         & \multicolumn{3}{c}{RNN} \\
     \cmidrule(l{2pt}r{2pt}){2-4} \cmidrule(l{2pt}r{2pt}){5-7} \cmidrule(l{2pt}r{2pt}){8-10} \cmidrule(l{2pt}r{2pt}){11-13}
   Metric & basic & full & strict & basic & full  &strict & basic & full & strict & basic & full & strict \\
   \midrule
\multicolumn{13}{l}{{\bfseries CDCP dataset}} \\
\smallspace   
             Average &      47.4  &      47.3  &      47.9  &      40.8  &      38.0  &      38.0  &      48.1  &      49.3  & {\bf 50.0} &      43.5  &      33.5  &      38.2  \\
\smallspace   
             {\Link} \cnt{272} &      22.0  &      21.9  &      23.8  &       9.9  &      12.8  &      12.8  &      24.7  &      25.1  & {\bf 26.7} &      14.4  &      14.6  &      10.5  \\
\smallspace   
             {\Prop} &      72.7  &      72.7  &      72.0  &      71.8  &      63.2  &      63.2  &      71.6  & {\bf 73.5} &      73.2  &      72.7  &      52.4  &      65.9  \\
{\enskip}{\typevalue} \cnt{491}&      75.3  &      75.3  &      74.4  &      74.1  &      74.8  &      74.8  &      73.4  &      75.7  &      76.4  &      73.7  &      73.1  &      69.7  \\
{\enskip}{\typepolicy} \cnt{153}&      78.7  &      78.7  &      78.5  &      74.3  &      72.2  &      72.2  &      72.3  &      77.3  &      76.8  &      73.9  &      74.4  &      76.8  \\
{\enskip}{\typetestimony} \cnt{204}&      70.3  &      70.3  &      68.6  &      74.6  &      71.8  &      71.8  &      69.8  &      71.7  &      71.5  &      74.2  &      72.3  &      75.8  \\
{\enskip}{\typefact} \cnt{124}&      39.2  &      39.2  &      38.3  &      35.8  &      30.5  &      30.5  &      42.4  &      42.5  &      41.3  &      41.5  &      42.2  &      40.5  \\
{\enskip}{\typereference} \cnt{1}&      100.0  &     100.0  &     100.0  &     100.0  &      66.7  &      66.7  &     100.0  &     100.0  &     100.0  &     100.0  &       0.0  &      66.7  \\

\midspace   
\multicolumn{13}{l}{{\bfseries UKP dataset}} \\
\smallspace
             Average &      64.7  &      66.6  &      66.5  &      58.7  &      57.4  &      58.7  &      67.1  & {\bf 68.9} &      67.1  &      59.0  &      63.6  &      64.7  \\
\smallspace
             {\Link} \cnt{809} &      55.8  &      59.7  & {\bf 60.3} &      44.8  &      43.8  &      44.0  &      56.9  &      60.1  &      56.9  &      44.1  &      50.4  &      50.1  \\
\smallspace
             {\Prop} &      73.5  &      73.5  &      72.6  &      72.6  &      70.9  &      73.3  &      77.2  &      77.6  &      77.3  &      74.0  &      76.9  & {\bf 79.3} \\
 {\enskip}{\typemc} \cnt{153}&      76.7  &      76.7  &      77.6  &      81.4  &      75.1  &      81.3  &      77.0  &      78.2  &      80.0  &      83.6  &      84.6  &      88.3  \\
 {\enskip}{\typeclaim} \cnt{304}&      55.4  &      55.4  &      52.0  &      51.7  &      52.7  &      53.5  &      64.3  &      64.5  &      62.8  &      53.2  &      60.2  &      62.0  \\
 {\enskip}{\typepremise} \cnt{809}&      88.4  &      88.4  &      88.3  &      84.8  &      84.8  &      85.2  &      90.3  &      90.2  &      89.2  &      85.0  &      85.9  &      87.6  \\

    \bottomrule \end{tabular} 
    
    \caption{Test set $F_1$ scores for {\link} and {\prop} classification, as
        well as their average, on the two datasets.  The number of test
        instances is shown in parentheses; best scores on overall tasks are in
        bold.
    } \label{tab:allscores} \end{table*}

We evaluate our proposed models on both datasets. For model selection and
development we used k-fold cross-validation at document level: on CDCP we set
$k=3$ to avoid small validation folds, while on UKP we follow \newcite{stabcoli}
setting $k=5$.  We compare our proposed structured learning systems (the linear
structured SVM and the structured RNN) to the corresponding baseline versions.
We organize our experiments in three incremental variants of our factor graph:
{\em basic}, {\em full}, and {\em strict}, each with the following components:%
\footnote{ Components are described in \secref{learn}.  The baselines {\em with
    inference} support only unaries and factors with no parameters, as indicated
    in the last column.
}

\vspace{4pt}
{
\centering \footnotesize
\begin{tabular}{l l l l l }
\toprule
component & basic & full & strict & (baseline) \\
\midrule
unaries & \checkmark & \checkmark & \checkmark & \checkmark \\
compat. factors & \checkmark & \checkmark & \checkmark & \\
compat. features &        & \checkmark & \checkmark & \\
higher-order & & \checkmark & \checkmark & \\
{\link} structure & & \checkmark & \checkmark & \checkmark\\
strict constraints & & & \checkmark & \checkmark \\
\bottomrule
\end{tabular}
}

\vspace{4pt}

\noindent Following \newcite{stabcoli}, we compute $F_1$ scores at {\prop} and
{\link} level, and also report their average as a summary of overall
performance.\footnote{For {\link} $F_1$ scores,
however, we find it more intuitive to only consider retrieval of positive links
rather than macro-averaged two-class scores.}
The results of a single prediction run on the test set are displayed in
\tabref{allscores}.  The overall trend is that training using a structured
objective is better than the baseline models, even when structured inference is
applied on the baseline predictions.  On UKP, for link prediction, the linear
baseline can reach good performance when using inference, similar to the
approach of \newcite{stabcoli}, but the improvement in {\prop} prediction leads
to higher overall $F_1$ for the structured models. Meanwhile, on the more
difficult CDCP setting, performing inference on the baseline output is not
competitive.  While feature engineering still outperforms our RNN model, we find
that RNNs shine on proposition classification, especially on UKP, and that
structured training can make them competitive, reducing their observed lag on
{\link} prediction \cite{arzoo}, possibly through mitigating class imbalance.
\subsection{Discussion and analysis}
\label{sec:erroranalysis}

\begin{figure*}[ht]
\centering\includegraphics[width=0.99\textwidth]{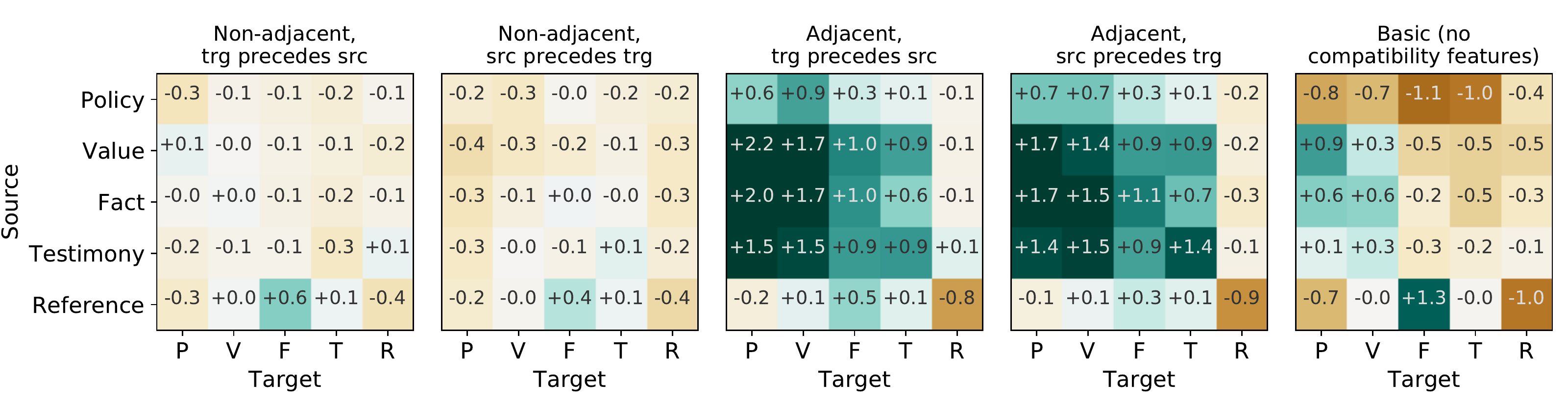}
\caption{Learned conditional log-odds $\log \frac{p(\mathtt{on} | \cdot )}
    {p(\mathtt{off}|\cdot)}$, given the 
source and target {\prop} types and compatibility feature settings.
First four figures correspond to the four possible
settings of the compatibility features in the full structured SVM model.
For comparison, the rightmost figure shows the same parameters in the basic
structured SVM model, which does not use compatibility features.}
\label{fig:compat}
\end{figure*}

\begin{figure}[t]
\centering\includegraphics[width=0.49\textwidth]{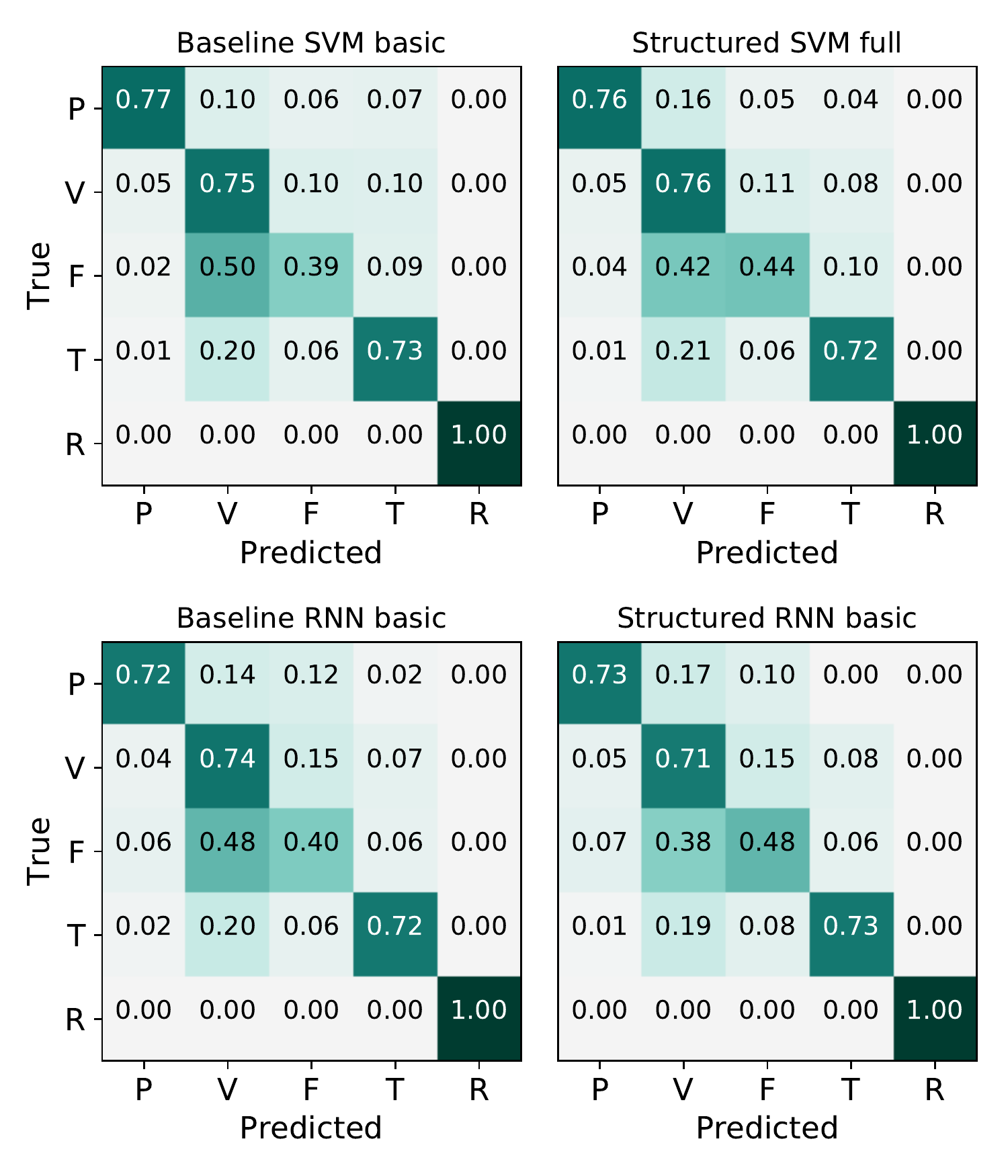}
\caption{Normalized confusion matrices for {\prop} type classification.}
\label{fig:confusion}
\end{figure}

\paragraph{Contribution of compatibility features.}  The compatibility factor in
our model can be visualized as conditional odds ratios given the source and
target {\prop} types. Since there are only four possible configurations of the
compatibility features, we can plot all cases in \figref{compat}, alongside the
basic model.  Not using compatibility features, the basic model can only learn
whether certain configurations are more likely than others (e.g. a
{\typereference} supporting another {\typereference} is unlikely, while a
{\typereference} supporting a {\typefact} is more likely; essentially a soft
version of our domain-specific strict constraints. The full model with
compatibility features is finer grained, capturing, for example, that {\links}
from {\typereference} to {\typefact} are more likely when the reference comes
{\em after}, or that {\links} from {\typevalue} to {\typepolicy} are extremely
likely only when the two are adjacent.

\paragraph{Proposition errors.} The confusion matrices in \figref{confusion}
reveal that the most common confusion is misclassifying {\typefact} as
{\typevalue}.  The strongest difference between the various models tested is
that the RNN-based models make this error less often. For instance, in the
{\prop}: \begin{quote}\footnotesize And the single most frequently used excuse
of any debtor is ``I didn't receive the letter/invoice/statement'' \end{quote}
the pronouns in the nested quote may be mistaken for subjectivity, 
leading to the structured SVMs predictions of {\typevalue} or
{\typetestimony}, while the basic structured RNN correctly classifies it as
{\typefact}.

\paragraph{Link errors.}
While structured inference certainly helps baselines by preventing invalid
structures such as cycles, it still depends on local decisions, losing to fully
structured training in cases where joint {\prop} and {\link} decisions are
needed. For instance, in the following conclusion of an UKP essay, the
annotators found no {\links}:

\begin{quote} \footnotesize In short, \qspopen{a}the individual should finance
    his or her education\qspclose{a}because \qspopen{b}it is a personal
    choice.\qspclose{b}Otherwise, \qspopen{c}it would cause too much cost from
    taxpayers and the government.\qspclose{c}
\end{quote}
Indeed, no reasons are provided, but baseline are misled by the connectives: the
SVM baseline outputs that $b$ and $c$ are {\typepremise}s supporting the
{\typeclaim} $a$. The full structured SVM combines the two tasks and correctly
recognizes the {\link} structure.

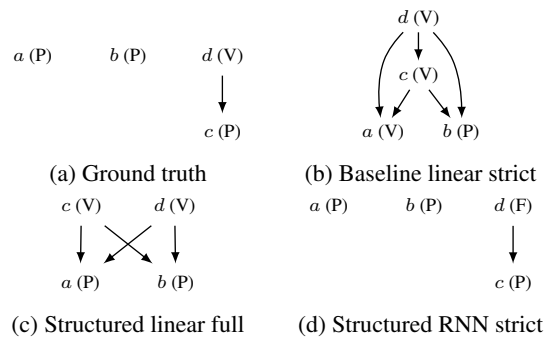
\begin{figure}[t]\begin{quote}
\footnotesize
\qspopen{a}I think the cost of education needs to be reduced
(...)
or repayment plans need to be income based.\qspclose{a}%
\qspopen{b}As far as consumer protection, legal aid needs to be made available,
affordable and effective,\qspclose{b}%
\qspopen{c}and consumers need to take time to really know their rights and stop
complaining about harassment\qspclose{c}%
\qspopen{d}because that's a completely different cause of action than
restitution.\qspclose{d}
\end{quote}
\scriptsize
\begin{minipage}[t]{.5\linewidth}
\centering

\begin{tikzpicture}[>=latex,line join=bevel,scale=0.4]
  \pgfsetlinewidth{0.5bp}
\pgfsetcolor{black}
  \draw [->] (212.5bp,71.697bp) .. controls (212.5bp,63.983bp) and (212.5bp,54.712bp)  .. (212.5bp,36.104bp);
\begin{scope}
  \definecolor{strokecol}{rgb}{0.0,0.0,0.0};
  \pgfsetstrokecolor{strokecol}
  \draw (35.5bp,90.0bp) node {$a$ (P)};
\end{scope}
\begin{scope}
  \definecolor{strokecol}{rgb}{0.0,0.0,0.0};
  \pgfsetstrokecolor{strokecol}
  \draw (212.5bp,18.0bp) node {$c$ (P)};
\end{scope}
\begin{scope}
  \definecolor{strokecol}{rgb}{0.0,0.0,0.0};
  \pgfsetstrokecolor{strokecol}
  \draw (124.5bp,90.0bp) node {$b$ (P)};
\end{scope}
\begin{scope}
  \definecolor{strokecol}{rgb}{0.0,0.0,0.0};
  \pgfsetstrokecolor{strokecol}
  \draw (212.5bp,90.0bp) node {$d$ (V)};
\end{scope}
\end{tikzpicture}

\subcaption{Ground truth}
\end{minipage}%
\begin{minipage}[t]{.5\linewidth}
\centering

\begin{tikzpicture}[>=latex,line join=bevel,scale=0.3]
  \pgfsetlinewidth{.5bp}
\pgfsetcolor{black}
  \draw [->] (78.0bp,143.7bp) .. controls (78.0bp,135.98bp) and (78.0bp,126.71bp)  .. (78.0bp,108.1bp);
  \draw [->] (67.124bp,71.697bp) .. controls (62.008bp,63.559bp) and (55.805bp,53.689bp)  .. (44.751bp,36.104bp);
  \draw [->] (91.101bp,71.697bp) .. controls (97.391bp,63.389bp) and (105.05bp,53.277bp)  .. (118.05bp,36.104bp);
  \draw [->] (59.024bp,143.81bp) .. controls (49.904bp,134.27bp) and (39.908bp,121.6bp)  .. (35.0bp,108.0bp) .. controls (27.885bp,88.288bp) and (28.028bp,64.484bp)  .. (30.914bp,36.323bp);
  \draw [->] (96.419bp,143.6bp) .. controls (105.39bp,133.99bp) and (115.41bp,121.33bp)  .. (121.0bp,108.0bp) .. controls (129.2bp,88.431bp) and (131.5bp,64.438bp)  .. (131.87bp,36.109bp);
\begin{scope}
  \definecolor{strokecol}{rgb}{0.0,0.0,0.0};
  \pgfsetstrokecolor{strokecol}
  \draw (34.0bp,18.0bp) node {$a$ (V)};
\end{scope}
\begin{scope}
  \definecolor{strokecol}{rgb}{0.0,0.0,0.0};
  \pgfsetstrokecolor{strokecol}
  \draw (78.0bp,90.0bp) node {$c$ (V)};
\end{scope}
\begin{scope}
  \definecolor{strokecol}{rgb}{0.0,0.0,0.0};
  \pgfsetstrokecolor{strokecol}
  \draw (131.0bp,18.0bp) node {$b$ (P)};
\end{scope}
\begin{scope}
  \definecolor{strokecol}{rgb}{0.0,0.0,0.0};
  \pgfsetstrokecolor{strokecol}
  \draw (78.0bp,162.0bp) node {$d$ (V)};
\end{scope}
\end{tikzpicture}

\subcaption{Baseline linear strict}
\end{minipage}

\begin{minipage}[t]{.5\linewidth}
\centering

\begin{tikzpicture}[>=latex,line join=bevel,scale=0.4]
  \pgfsetlinewidth{.5bp}
\pgfsetcolor{black}
  \draw [->] (36.253bp,71.697bp) .. controls (36.143bp,63.983bp) and (36.01bp,54.712bp)  .. (35.744bp,36.104bp);
  \draw [->] (58.253bp,71.697bp) .. controls (69.336bp,62.881bp) and (82.974bp,52.032bp)  .. (103.0bp,36.104bp);
  \draw [->] (101.75bp,71.697bp) .. controls (90.664bp,62.881bp) and (77.026bp,52.032bp)  .. (57.003bp,36.104bp);
  \draw [->] (123.75bp,71.697bp) .. controls (123.86bp,63.983bp) and (123.99bp,54.712bp)  .. (124.26bp,36.104bp);
\begin{scope}
  \definecolor{strokecol}{rgb}{0.0,0.0,0.0};
  \pgfsetstrokecolor{strokecol}
  \draw (35.5bp,18.0bp) node {$a$ (P)};
\end{scope}
\begin{scope}
  \definecolor{strokecol}{rgb}{0.0,0.0,0.0};
  \pgfsetstrokecolor{strokecol}
  \draw (36.5bp,90.0bp) node {$c$ (V)};
\end{scope}
\begin{scope}
  \definecolor{strokecol}{rgb}{0.0,0.0,0.0};
  \pgfsetstrokecolor{strokecol}
  \draw (124.5bp,18.0bp) node {$b$ (P)};
\end{scope}
\begin{scope}
  \definecolor{strokecol}{rgb}{0.0,0.0,0.0};
  \pgfsetstrokecolor{strokecol}
  \draw (123.5bp,90.0bp) node {$d$ (V)};
\end{scope}
\end{tikzpicture}

\subcaption{Structured linear full}
\end{minipage}%
\begin{minipage}[t]{.5\linewidth}
\centering

\begin{tikzpicture}[>=latex,line join=bevel,scale=0.4]
  \pgfsetlinewidth{.5bp}
\pgfsetcolor{black}
  \draw [->] (207.5bp,71.697bp) .. controls (207.5bp,63.983bp) and (207.5bp,54.712bp)  .. (207.5bp,36.104bp);
\begin{scope}
  \definecolor{strokecol}{rgb}{0.0,0.0,0.0};
  \pgfsetstrokecolor{strokecol}
  \draw (35.5bp,90.0bp) node {$a$ (P)};
\end{scope}
\begin{scope}
  \definecolor{strokecol}{rgb}{0.0,0.0,0.0};
  \pgfsetstrokecolor{strokecol}
  \draw (207.5bp,18.0bp) node {$c$ (P)};
\end{scope}
\begin{scope}
  \definecolor{strokecol}{rgb}{0.0,0.0,0.0};
  \pgfsetstrokecolor{strokecol}
  \draw (124.5bp,90.0bp) node {$b$ (P)};
\end{scope}
\begin{scope}
  \definecolor{strokecol}{rgb}{0.0,0.0,0.0};
  \pgfsetstrokecolor{strokecol}
  \draw (207.5bp,90.0bp) node {$d$ (F)};
\end{scope}
\end{tikzpicture}

\subcaption{Structured RNN strict}
\end{minipage}
\caption{Predictions on a CDCP comment where the structured
RNN outperforms the other models.}
\label{fig:error}
\end{figure}

Linear SVMs are still a very good baseline, but they tend to overgenerate
{\links} due to class imbalance, even if we use class weights during training.
Surprisingly, RNNs are at the opposite end, being extremely conservative, and
getting the highest precision among the models.  On CDCP, where the number of
true {\links} is 272, the linear baseline with strict inference predicts 796
{\links} with a precision of only 16\%, while the strict structured RNN only
predicts 52 {\links}, with 33\% precision; the example in \figref{error}
illustrates this.  In terms of higher-order structures, we find that using
higher-order factors increases precision, at a cost in recall. This is most
beneficial for the 856 co-parent structures in the UKP test set: the full
structured SVM has 53\% $F_1$, while the basic structured SVM and the basic
baseline get 47\% and 45\% respectively.  On CDCP, while higher-order factors
help, performance on siblings and co-parents is below 10\% $F_1$ score. This is
likely due to {\link} sparsity and suggests plenty of room for further
development.

\section{Conclusions and future work} \label{sec:conclusions} We introduce an
argumentation parsing model based on {\adq} relaxed inference in expressive
factor graphs, experimenting with both linear structured SVMs and structured
RNNs, parametrized with higher-order factors and {\link} structure constraints.
We demonstrate our model on a new argumentation mining dataset with more
permissive argument structure annotation. Our model also achieves
state-of-the-art {\link} prediction performance on the UKP essays dataset.
\paragraph{Future work.} \newcite{stabcoli} found polynomial kernels useful for
modeling feature interactions, but kernel structured SVMs scale poorly, we
intend to investigate alternate ways to capture feature interactions.  While we
focus on monological argumentation, our model could be extended to dialogs,
for which argumentation theory thoroughly motivates non-tree structures
\cite{asher}.

\section*{Acknowledgements}
We are grateful to
Andr\'{e} Martins,
Andreas M\"{u}ller,
Arzoo Katyiar,
Chenhao Tan,
Felix Wu,
Jack Hessel,
Justine Zhang,
Mathieu Blondel,
Tianze Shi,
Tobias Schnabel,
and the rest of the Cornell NLP seminar for extremely helpful discussions.
We thank the anonymous reviewers for their thorough and well-argued feedback. 

\bibliographystyle{acl_natbib}

\end{document}